\documentclass[sigconf]{acmart}
 \providecommand\BibTeX{{%
  Bib\TeX}}

\AtBeginDocument{%
  \providecommand\BibTeX{{%
    \normalfont B\kern-0.5em{\scshape i\kern-0.25em b}\kern-0.8em\TeX}}}

\copyrightyear{2024}
\acmYear{2024}
\setcopyright{rightsretained}
\acmConference[SIGIR '24]{Proceedings of the 47th International ACM SIGIR Conference on Research and Development in Information Retrieval}{July 14--18, 2024}{Washington, DC, USA}
\acmBooktitle{Proceedings of the 47th International ACM SIGIR Conference on Research and Development in Information Retrieval (SIGIR '24), July 14--18, 2024, Washington, DC, USA}
\acmDOI{10.1145/3626772.3657872}
\acmISBN{979-8-4007-0431-4/24/07}

\usepackage[ruled,linesnumbered]{algorithm2e}
\usepackage{multirow}
\usepackage{todonotes}
\usepackage{hyperref}
\usepackage{booktabs}
\usepackage{comment}
\usepackage{nicematrix}
\usepackage{tabularx}
\NiceMatrixOptions{ custom-line = { letter = ; , tikz = dashed } }

\makeatletter
\gdef\@copyrightpermission{
  \begin{minipage}{0.3\columnwidth}
   \href{https://creativecommons.org/licenses/by/4.0/}{\includegraphics[width=0.90\textwidth]{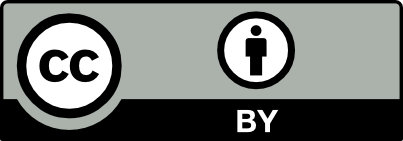}}
  \end{minipage}\hfill
  \begin{minipage}{0.7\columnwidth}
   \href{https://creativecommons.org/licenses/by/4.0/}{This work is licensed under a Creative Commons Attribution International 4.0 License.}
  \end{minipage}
  \vspace{5pt}
}
\makeatother

\begin{document}

\title{ACE-2005-PT: Corpus for Event Extraction in Portuguese}


\author{Luís Filipe Cunha}
\email{lfc@uminho.pt}
\orcid{0000-0003-1365-0080}
\affiliation{%
  \institution{FCUP-University of Porto \and INESC Technology and Science 
  \and
  \city{Porto}
  \country{Portugal}
  \and University of Minho - Braga, Portugal}
}

\author{Purificação Silvano}
\email{msilvano@letras.up.pt}
\orcid{0000-0001-8057-5338}
\affiliation{%
  \institution{FLUP-University of Porto, \and CLUP}
  \city{Porto}
  \country{Portugal}
}

\author{Ricardo Campos}
\email{ricardo.campos@ubi.pt}
\orcid{0000-0002-8767-8126}
\affiliation{%
  \institution{University of Beira Interior \and   \city{Covilhã}
  \country{Portugal} \and INESC Technology and Science\and   \city{Porto} \country{Portugal} \and
  Ci2 - Smart Cities Research Center - IP Tomar}
  \city{Tomar}
  \country{Portugal}
}

\author{Alípio Jorge}
\email{amjorge@fc.up.pt}
\orcid{0000-0002-5475-1382}
\affiliation{%
  \institution{FCUP-University of Porto
\and INESC Technology and Science}
  \city{Porto}
  \country{Portugal}
}


\begin{abstract}
Event extraction is an NLP task that commonly involves identifying the central word (trigger) for an event and its associated arguments in text. ACE-2005 is widely recognised as the standard corpus in this field. While other corpora, like PropBank, primarily focus on annotating predicate-argument structure, ACE-2005 provides comprehensive information about the overall event structure and semantics. However, its limited language coverage restricts its usability. This paper introduces ACE-2005-PT, a corpus created by translating ACE-2005 into Portuguese, with European and Brazilian variants. To speed up the process of obtaining ACE-2005-PT, we rely on automatic translators. This, however, poses some challenges related to automatically identifying the correct alignments between multi-word annotations in the original text and in the corresponding translated sentence. To achieve this, we developed an alignment pipeline that incorporates several alignment techniques: lemmatization, fuzzy matching, synonym matching,  multiple translations and a BERT-based word aligner. To measure the alignment effectiveness, a subset of annotations from the ACE-2005-PT corpus was manually aligned by a linguist expert. This subset was then compared against our pipeline results which achieved exact and relaxed match scores of 70.55\% and 87.55\% respectively. As a result, we successfully generated a Portuguese version of the ACE-2005 corpus, which has been accepted for publication by LDC.
\end{abstract}





\maketitle

\section{Introduction}

Event extraction is a crucial Information Extraction task, that has witnessed a growing interest among researchers in recent years \cite{li2022survey}. However, this task can pose significant challenges due to its inherent complexity and the diverse ways in which events can be expressed in natural language \cite{li-etal-2013-joint}. Despite these challenges, several works \cite{nguyen-etal-2022-joint, nguyen-etal-2021-cross, lu-etal-2021-text2event,BALALI2020106492} have demonstrated significant progress in this field, achieving promising results.

One of the most widely used corpora in event extraction research is the ACE-2005 (Automatic Content Extraction) corpus \cite{doddington-etal-2004-automatic}, which has become a gold standard in the field \cite{jurafsky:2023}. This corpus provides a comprehensive collection of annotated events, offering a valuable benchmark for evaluating the effectiveness of event extraction systems. Unfortunately, it is only available in English, Chinese and Arabic, but not in other languages, such as Portuguese. 

In order to bridge this gap and promote event extraction for the Portuguese language, we introduce ACE-2005-PT (Automatic Content Extraction for Portuguese), an automated translation of the ACE-2005 corpus that has been partially verified by a linguist expert. To generate this corpus, we developed a translation and annotation alignment pipeline that combines machine translation and several text alignment techniques such as fuzzy string matching, lemmatization, synonym matching, multiple translations and a word aligner based on Transformers \cite{vaswani2017attention}. By leveraging this pipeline, we have successfully translated the ACE-2005 corpus into Portuguese, including two variants: European Portuguese and Brazilian Portuguese. Then, a subset of ACE-2005-PT annotations was manually aligned by a linguist to validate the pipeline's effectiveness. 

This corpus has already been accepted for publication by the Linguistic Data Consortium (LDC) and will be available for event extraction-related tasks. The main contributions of this paper are:

\begin{itemize}
    \item An automatic translation and alignment pipeline that can be used to translate not only ACE-2005 but other corpora into different languages. We made the code available\footnote{\url{https://github.com/LIAAD/ACE-2005-Translation-and-Alignment-Pipeline}} for reproducibility purposes. 
    \item A Portuguese-translated version of the ACE-2005 corpus that can be used to enhance Event Extraction in two Portuguese variants: European and Brazilian.
\end{itemize}

In the following sections, we present a brief description of ACE-2005 and the methodology used to translate it into Portuguese.

\section{ACE-2005 Corpus}\label{sec:data}

The ACE-2005 corpus is composed of several textual documents from various sources, such as newswires, online journals, broadcast transcripts, discussion forums and conversational telephone speech. Each document is provided with annotations of events, which consist of event triggers and their corresponding event arguments \cite{aceguidelines}. Event triggers represent the terms that indicate the occurrence of an event and play a crucial role in event extraction. Each event trigger is associated with a specific event type resulting in 33 event types in the ACE-2005 corpus. Event arguments describe entities, temporal expressions or values that serve as participants or attributes of the events. Each event argument is associated with a semantic role that represents its relationship within the event, such as the agent that performs an action and the time or location of the event. For each event occurrence, ACE-2005 annotates event arguments by linking them to their respective event trigger. 

Consider the following sentence, which illustrates the process of extracting an event trigger and its corresponding arguments. 

\begin{center}
Marie Curie was born in Warsaw on November 7, 1867.
\end{center}
In this example, the word "born" corresponds to a trigger with type \texttt{Life:Be-Born}. Regarding the event arguments, we have "Marie Curie", "Warsaw" and "November 7, 1867" with roles \texttt{Person}, \texttt{Place} and \texttt{Time} respectively.


\section{ACE-2005 Translation to Portuguese}

In this section, we introduce the translation process and annotation alignment pipeline used to generate the ACE-2005-PT corpus and present statistics associated with this methodology. Following that, we introduce the methods used to evaluate our pipeline effectiveness and the obtained results. 

\subsection{Translation and Annotation Alignment}\label{sec:translate}

The initial step in the process of producing the ACE-2005 corpus consisted of performing automatic translation of the documents. Specifically, we used Google Translator\footnote{https://cloud.google.com/translate/docs/reference/rest} for Brazilian Portuguese, while for European Portuguese, we relied on DeepL translator \footnote{https://www.deepl.com}, as Google Translator does not support translation into this language variant.

After translation, we have an original text with annotated terms and we want to transfer these annotations to the translated text, in a process called alignment (of annotations). In the simplest case, we translate the annotated term and its translation becomes annotated in the translated text. However, such a direct approach works only half of the time. For instance, in the sentence ``The soldiers were ordered to fire their weapons", ACE-2005 states that the trigger ``fire" should be annotated. However, this sentence is translated to "\textit{Os soldados receberam ordens para disparar as suas armas}" where the word ``fire" is translated to ``\textit{incêndio}" (fire as a noun) in isolation and to ``\textit{disparar}" (fire as a verb) in context.

We employed a commonly used pre-processing approach\footnote{https://github.com/nlpcl-lab/ace2005-preprocessing} of the ACE-2005 corpus, which performed sentence tokenization on the documents and assigned each event annotation to its respective sentence. Subsequently, we automatically translated each source (${src}$) sentence $s_{src}$, along with its corresponding triggers $t_{src}$ and arguments $a_{src}$. In case a mismatch occurs, these translations (${trans}$) result in annotations $a_{trans}$ and $t_{trans}$ that are not contained in the sentences $s_{trans}$. For these cases, we applied an alignment procedure (Algorithm \ref{alg:align_pipeline}) to identify the correct span offsets of $a_{trans}$ and $t_{trans}$ within $s_{trans}$. For the sake of simplicity, we only describe the event argument alignment process as the trigger alignment procedure was similar. The alignment pipeline is composed of four components: lemmatization, multiple translations, a BERT-based word aligner and fuzzy string similarity.

\RestyleAlgo{ruled}
\SetKwComment{Comment}{/*}{*/}
 \begin{algorithm}
            \SetAlgoLined
            
            \caption{Translation and Alignment Pipeline.}
            \label{alg:align_pipeline}
                \For{$s_{src}$ in sentences}{ 
        $s_{trans} \gets get\ sentence\ translation$\;
        \For{$a_{src}$ in event arguments}{
        
            $a_{trans} \gets get\ argument\ translation$\;
            \lIf{$a_{trans}$ in $s_{trans}$}{
              \Return $a_{trans}$
              }
                    $match \gets LemmaMatch(a_{lem},s_{lem})$\;
                    \lIf{$match$}{
                      \Return $match$
                      }

                    $match \gets MultipleTranslation(a_{src},s_{trans})$\;
                    \lIf{$match$}{
                      \Return $match$
                      }
                      
                         $match \gets WordAligner(s_{src},s_{trans},a_{src)}$\;
                        \lIf{$match$}{
                          \Return $match$
                          }
            \Return $FuzzyMatch(a_{trans},s_{trans})$\;
            }
    }
\end{algorithm}

The initial step of the alignment pipeline starts by lemmatizing both $s_{trans}$ and $a_{trans}$, resulting in $s_{trans}^{lem}$ and $a_{trans}^{lem}$, as shown in Algorithm \ref{alg:align_Lemma}. Next, we verify whether the sequence of tokens in $a_{trans}^{lem}$ occurs in $s_{trans}^{lem}$. By transforming each token of $s_{trans}$ and $a_{trans}$ into their fundamental word forms, we reduce the potential variations for each word and increase the likelihood of finding a match. 

\begin{algorithm}[h]
    \SetAlgoLined

\caption{LemmaMatch function.}\label{alg:align_Lemma}
\SetKwProg{Def}{def}{:}{}
\Def{getLemmaMatch$(s_{trans},\ a_{trans})$}{
    $a_{trans}^{lem} \gets get\ a_{trans}\ lemma\ tokens$\;
    $s_{trans}^{lem} \gets get\ s_{trans}\ lemma\ tokens $\;
    \uIf{$a_{trans}^{lem}\ in \ s_{trans}^{lem}$ }{
        $start, end \gets find\_span(s_{trans}^{lem},\ a_{trans}^{lem})$\;
      \Return $a_{trans} \gets s_{trans}[start,end]$\;
    }
    }
\end{algorithm}

The next element of the pipeline consists of using multiple translations of words corresponding to annotated terms. In particular, we used the Microsoft Dictionary Lookup API\footnote{https://learn.microsoft.com/en-us/azure/ai-services/translator/reference/v3-0-dictionary-lookup} which provides alternative translations for words and idiomatic expressions to generate alternative translations of $a_{trans}$ and $t_{trans}$, aiming to locate them in $s_{trans}$. This method was also combined with the previous one by performing lemmatization on the alternative translation and trying to match it within the $s_{trans}$.

Following these methods, we used a parallel corpus word aligner adapted from the word aligner proposed by Dou et al. (2021) \cite{Dou2021}. We retrieved the embedding matrices of $s_{src}$ and $s_{trans}$ from the BERT-Multilingual model (mBERT) \cite{DBLP:journals/corr/abs-1810-04805} and computed their product. Subsequently, we applied a softmax function to compute probabilities for the association between tokens in $s_{src}$ and $s_{trans}$.

\begin{figure}[h]

    \centering
    \includegraphics[width=8cm]{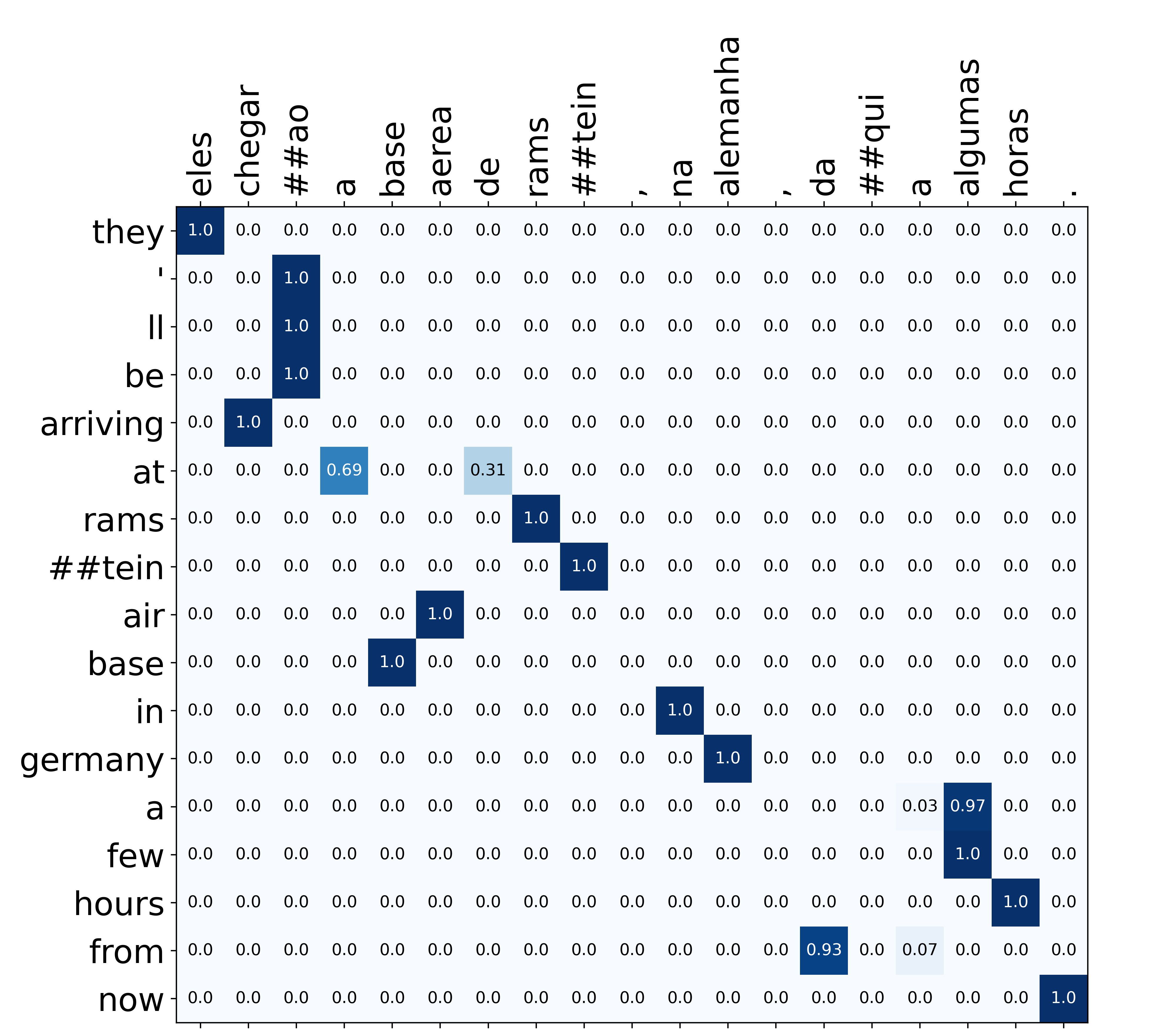}
    \caption{Alignment matrix from mBERT.}
  \label{fig:aligns}
  
\end{figure}

Figure \ref{fig:aligns} presents an example of the embeddings-driven word association matrix with an English sentence $s_{src}$ on the vertical axis and its corresponding Portuguese translation $s_{trans}$ on the horizontal axis. 
With the word alignments between $s_{src}$ and $s_{trans}$ computed, we proceeded to find the correct $a_{trans}$ span offset. For that, we used an approach represented in Algorithm \ref{alg:align_wordaligner}, which is similar to the one described by Carrino et al. (2020)\cite{Carrino2020}, with the inclusion of a token size verification. First, we calculate the start ($a_{src}^{start}$) and end ($a_{src}^{end}$) positions of $a_{src}$ in $s_{src}$. Next, we use the word alignments computed in $align(s_{src}, s_{trans})$ to retrieve the corresponding start and end positions of $a_{trans}$ in $s_{trans}$, retrieving $a_{trans}^{start}$ and $a_{trans}^{end}$ respectively. Finally, we determine the corresponding $a_{trans}$ by using its span offset as indexes in $s_{trans}$. Then we verify the length of the aligned annotation by comparing it with the original annotation length to avoid returning disproportionately large annotations. (The motivation for this is further explained in Section \ref{sec:limitations}).

 \begin{algorithm}
      \SetAlgoLined
      \caption{Word Aligner Algorithm}
      \label{alg:align_wordaligner}
      \SetKwProg{Def}{def}{:}{}
      \Def{WordAligner$(s_{src},\  s_{trans},\ a_{src})$}{
        $aligns \gets align(s_{src},s_{trans})$ \Comment*[r]{mBERT}
        
        $(a_{\text{src}}^{\text{start}},...,a_{\text{src}}^{\text{end}}) \gets $ {\it get} $a_{src}$\ {\it span offsets}\;
        
        ($a_{\text{trans}}^{\text{start}}, ..., a_{\text{trans}}^{\text{end}}$) $\gets a_{\text{trans}}$ \textit{from aligns};
        
        $a_{\text{trans}}^{'\text{start}} \gets \min(a_{\text{trans}}^{\text{start}},a_{\text{trans}}^{\text{end}})$\;
        
        $a_{\text{trans}}^{'\text{end}} \gets \max(a_{\text{trans}}^{\text{start}},a_{\text{trans}}^{\text{end}})$\;
        
        $a_{\text{trans}} \gets s_{\text{trans}}a_{\text{trans}}^{'\text{start}},a_{\text{trans}}^{'\text{end}}$\;

        \If{$SizeSafeguard(a_{\text{trans}},a_{\text{src}})$}{
          \Return $a_{\text{trans}}$
          }
      }
    \end{algorithm}

The last component of the pipeline uses Fuzzy string similarity. We calculate the similarity between $a_{trans}$ and substrings of $s_{trans}$. To generate these substrings, we use a sliding window to compute all substrings of $s_{trans}$ with length $window$ equal to the number of tokens in $a_{trans}$. In order to determine the most similar string between $a_{trans}$ and the generated substrings, we use fuzzy matching algorithms such as the Levenshtein distance \cite{levenshtein1965} and Gestalt pattern matching \cite{ratcliff1988pattern}.

In the case of trigger alignment, where triggers generally consist of a single word, we did not use the fuzzy match component. This decision was based on the limitations associated with character-level similarity techniques, which are further elaborated in Section \ref{sec:limitations}. Instead, we used synonyms of $t_{trans}$, trying to establish a correspondence between $t_{trans}$ and $s_{trans}$.

\subsection{Statistics by pipeline component}

Using the alignment pipeline, we produced ACE-2005-PT, featuring 16,260 sentences and 14,886 annotations (triggers and arguments), mirroring the original corpus. In this section, we focus, for the sake of simplicity, on the European corpus, as the results were similar in the Brazilian variant. Upon translation, we were able to find 2,721 triggers (51.9\%) and 5,127 arguments (53.1\%) in their respective translated sentences by using substring matching between the annotations and the translated text. This indicates that approximately half of the ACE-2005-PT annotations required alignment.


\begin{table}[h]
\caption{ACE-2005-PT alignment statistics. Number of alignments by component. (SMatch - string match; MTrans - multiple translation; WAligner - word aligner) }
\label{tab:pipeline_stats}
\centering
\begin{NiceTabular}
{@{}ccccccccc@{}}
\toprule
\multicolumn{1}{c}{\multirow{2}{*}{\textbf{Method}}}                    & \multicolumn{4}{c}{\textbf{Trigger  }}           & \multicolumn{4}{c}{\textbf{Argument }}   \\  
\multicolumn{1}{c}{}                                                   & \multicolumn{1}{c}{Train} & \multicolumn{1}{c}{Dev} & \multicolumn{1}{c}{Test} & \multicolumn{1}{c}{Total} &\multicolumn{1}{c}{Train}     & \multicolumn{1}{c}{Dev}     & \multicolumn{1}{c}{Test}  & \multicolumn{1}{c}{Total} \\ \midrule

SMatch & 2,289        &    222     &      \multicolumn{1}{c}{210}  &  2,721 &  4,109        & 503   & 515 & 5,127     \\

Lemma & 473        &    56     &      \multicolumn{1}{c}{38}  &  567 &  1,235        & 152   & 136 & 1,523                \\

MTrans & 947   &     144    &       \multicolumn{1}{c}{125}   & 1,216      & 568              & 53   & 52     & 673           \\

Synonym & 16   &     2    &       \multicolumn{1}{c}{1}    & 19    & -               & -   & -      & -          \\ 

WAligner & 587    &     66     &     \multicolumn{1}{c}{47}  &  700    & 1,675               & 198   & 168       & 2,041         \\ 

Fuzzy & -          &      -    &     \multicolumn{1}{c}{-}     &    -     & 152            & 25   & 16        & 193        \\ 

Manual &  12         &     2     &     \multicolumn{1}{c}{1}    & 15    & 84               & 2   & 5            & 91    \\ 
                          


\midrule
Total  & 4,324        &    492     &      \multicolumn{1}{c}{422}  &  5,238 &  7,823        & 933   & 892 & 9,648     \\

 \bottomrule

\CodeAfter 
  \tikz \draw [dashed] (2-|5) -- (11-|5) 
                         
                       (2-|9) -- (11-|9)
                        ;
  
  \tikz \draw          (1-|2) -- (11-|2)
                        (2-|2) -- (2-|11)
                        (1-|6) -- (11-|6);

\end{NiceTabular}

\end{table}

Table \ref{tab:pipeline_stats} provides the number of alignments performed by each component of our alignment pipeline for each data split. Notably, the pipeline works sequentially, meaning that annotations aligned by earlier methods are not addressed again by subsequent pipeline elements. 
To select the best component order in the pipeline we experimented with all the permutations between the components and calculated the corresponding alignment results using a manually aligned corpus that is further introduced in Section \ref{sec:Evaluation}. The
obtained results confirmed that the order that presents the best results is the one presented in Table \ref{tab:pipeline_stats} (top to bottom).    

Despite some entries couldn't be automatically aligned using our pipeline, their number was marginal compared to the total number of annotations. After using the alignment pipeline, we conducted a set of 106 manual alignments, corresponding to the "Manual" method in Table \ref{tab:pipeline_stats}.

\subsection{Evaluation}\label{sec:Evaluation}

To measure the effectiveness of the alignment pipeline, manual alignments were conducted on the entire ACE-2005-PT test set, which includes 1,310 annotations (triggers and arguments). These alignments were performed by an expert linguist to ensure high-quality annotations, following the same annotation guidelines of the original ACE-2005 corpus \cite{aceguidelines}. These annotations were then compared with the automatic alignment performed by our pipeline. During this comparison, we used two different metrics: the exact match between the pipeline-generated alignments and the manual ones (strict), and the F1 score between the tokens found by the pipeline and the ones that were manually aligned (relaxed). Resorting to both evaluation metrics allow us to more accurately assess the model's effectiveness. 

\subsection{Results}

Table \ref{tab:pipeline_results} shows the alignment validation results of our pipeline when compared with the manually annotated test split of ACE-2005-PT.

\begin{table}[h]
\caption{Evaluation Results by pipeline component.}
\label{tab:pipeline_results}
  \centering
\begin{NiceTabular}{@{}ccccccccccc@{}}
    \toprule
\multicolumn{1}{c}{\multirow{2}{*}{\textbf{Method}}}                    & \multicolumn{2}{c}{\textbf{Trigger  }}           & \multicolumn{2}{c}{\textbf{Argument }} & \multicolumn{2}{c}{\textbf{All}}   \\  
\multicolumn{1}{c}{}                                                   & \multicolumn{1}{c}{Relaxed} & \multicolumn{1}{c}{Exact} & \multicolumn{1}{c}{Relaxed}  &\multicolumn{1}{c}{Exact}  & \multicolumn{1}{c}{Relaxed} & \multicolumn{1}{c}{Exact}  \\ \midrule

    SMatch & 99.29 & 99.05 & 92.21 & 76.36 &  94.25  &  82.92 \\
    Lemma   & 93.86 & 92.11 & 74.06 & 20.74 & 78.41    & 36.42   \\
    MTrans   & 93.73 & 90.40 & 77.04 & 46.15 & 88.83    & 77.40   \\
    Synonym  & 1.0 & 1.0 & - & - & 1.0      & 1.0    \\
    WAligner & 63.54 & 58.33 & 68.26 & 28.57 & 67.22    & 35.19   \\
    Fuzzy   & - & - & 92.86 & 56.25 & 92.86  &    56.25    \\
    \midrule
    Pipeline   & 93.11& 91.23 & 85.16 & 60.76 & 87.77 & 70.55 \\
    \bottomrule

  \label{tab:evaluation}

  \CodeAfter 

  \tikz \draw          (1-|2) -- (10-|2)
                        (2-|2) -- (2-|8)
                        (1-|4) -- (10-|4)
                        (1-|6) -- (10-|6);

\end{NiceTabular}
\end{table}

In this table, we can analyse the relaxed and exact match between the pipeline-generated alignments and the gold annotations performed by the linguistic expert. The results are presented for each alignment method, along with the overall performance of the pipeline. The pipeline achieved a Relaxed score of 87.72\% and an exact score of 70.55\%. As expected, the flexible match obtained a higher result value than the exact match. Looking at the overall performance, there is a 17.22\% difference between both metrics, which shows that, despite our pipeline getting most of the alignments in the correct form, it still presents difficulties in reproducing the exact alignments performed by an expert.

Looking at the annotation types (triggers and arguments), our pipeline demonstrates more difficulties when aligning arguments (85.16\% relaxed and 60.76\% exact) compared to triggers (93.11\% relaxed and 91.23\% exact). This is expected since a trigger is usually composed of one word, while arguments may contain several words, making it harder to identify the correct alignment. Another observation to be made is that in the case of trigger alignment, the results obtained for both relaxed and exact scores are similar (93.11\% relaxed and 91.23\% exact). On the other hand, one can observe, in the argument alignments, a higher difference between the relaxed and the exact score, with a score of 85.15\% and 60.76\% respectively. Again, this indicates that aligning arguments is a harder task for our pipeline.

Finally, looking at the aligning modules individually, one can observe that the string match of the translated annotation with the translated text is the method that yielded the highest evaluation scores, for both triggers and arguments. Both the Lemmatization (Lemma) and the Multiple Word Translation (MTrans) methods revealed alignment scores above 90\% on the trigger alignment. As for the argument alignment, the Fuzzy string match and MTrans were the ones that achieved better alignment scores. The limitations of each alignment technique are further explored in Section \ref{sec:limitations}.

The quality of these alignments can further be assessed by observing the Event Extraction task results of models that were trained using the ACE-2005-PT corpus. In Cunha et. al (2023) \cite{cunha2023} this corpus was used to train Question Answering models to perform Event Extraction for Portuguese. The model trained with ACE-2005-PT is available for direct scrutiny online\footnote{https://hf.co/spaces/lfcc/Event-Extractor}.

Table \ref{tab:results_EE} presents F1-scores for trigger extraction and argument extraction on both the ACE-2005 corpus in English (original) and our Portuguese version presented in this paper, ACE-2005-PT. 

\begin{table}[h]

\caption{Event Extraction results on ACE-2005 dataset.}
\label{tab:results_EE}
\begin{tabularx}{8.5cm}{XXX}
\toprule
\multicolumn{1}{c}{\textbf{Model}} & \multicolumn{1}{r}{\textbf{Triggers F1}} &  \multicolumn{1}{r}{\textbf{Arguments F1}} \\ 
 \midrule

\multicolumn{3}{c}{\textbf{English ACE-2005}} \\ \midrule
\multicolumn{1}{l|}{JRNN 2016 \cite{nguyen-etal-2016-joint-event}} & \multicolumn{1}{c}{69.3} & \multicolumn{1}{c}{55.4} \\
\multicolumn{1}{l|}{BERT\_QA\_Arg 2020 \cite{Du2020}} & \multicolumn{1}{c}{\textbf{72.4}} & \multicolumn{1}{c}{\textbf{53.3}} \\
\multicolumn{1}{l|}{OneIE 2020 \cite{lin-etal-2020-joint}} & \multicolumn{1}{c}{74.7} & \multicolumn{1}{c}{56.8} \\
\multicolumn{1}{l|}{Text2Event 2021 \cite{lu-etal-2021-text2event}} & \multicolumn{1}{c}{71.9} & \multicolumn{1}{c}{53.8} \\
\multicolumn{1}{l|}{FourIE 2021 \cite{nguyen-etal-2021-cross}} & \multicolumn{1}{c}{75.4} & \multicolumn{1}{c}{58.0} \\
\multicolumn{1}{l|}{GraphIE 2022 \cite{nguyen-etal-2022-joint}} & \multicolumn{1}{c}{75.7} & \multicolumn{1}{c}{59.4} \\ \midrule

\multicolumn{3}{c}{\textbf{Portuguese ACE-2005}} \\ \midrule
\multicolumn{1}{l|}{BERT-ACE05PT \cite{cunha2023} (ours)} & \multicolumn{1}{c}{\textbf{64.4}} & \multicolumn{1}{c}{\textbf{46.7}} \\ \bottomrule
\end{tabularx}
\end{table}

Since the Portuguese model consists of a QA model it should primarily be compared with the BERT\_QA\_Arg model in Table \ref{tab:results_EE} due to the similarities in their model architectures. Looking at the table, one can observe that the results for the Portuguese model (64\% F1-score for Trigger Extraction and 46\% F1-score for Argument Extraction) are lower when compared to the results of the English models.  However, since the language used to train and evaluate these models is different, a direct comparison might be inaccurate due to divergences in language vocabulary, such as idiomatic expressions, and other cultural differences between languages.

\section{Error Analysis and Limitations}\label{sec:limitations}

Firstly, it is important to acknowledge that automatic translation is prone to translation errors which can lead to inaccuracies and a loss of the intended meaning in the translated texts. 

Regarding annotation alignments, some errors were found. For instance, in non-null subject languages like English, it is obligatory to include the subject in sentence construction, while in Portuguese, the subject can be implicit \cite{barbosa2005null}. Consider the next sentence as an example: 

\begin{center}

"We have been under heavy fire in the Middle East" 

\end{center}

In this example, the word "fire" is marked as an event trigger of type \texttt{Conflict:Attack}, and "We" is an event argument with the role \texttt{Target}. However, in the Portuguese translation, "We have" was translated to "\textit{Temos}" (the conjugated verb "have" in the first person plural). As a result, the subject became implicit, omitting the argument "We". During the validation process on the test split of the ACE-2005-PT corpus, we detected 41 errors in the alignment of arguments related to this phenomenon. These errors account for 11.71\% of the total argument alignment errors.

Another identified issue consists of our pipeline's inability to accurately identify the sentences' syntax structure. According to the ACE-2005 annotation guidelines \cite{aceguidelines}, when annotating a noun phrase (NP) with a determiner, the latter should be included in the argument. For instance, in the previous example, the NP "\textit{the Middle East}", which is an event argument with the role \texttt{Place}, comprises the determiner "\textit{the}". In this case, the NP is part of a prepositional phrase (PP) headed by the preposition "\textit{in}", but the preposition is not part of the argument. In the context of the Portuguese language, in examples like this one, the preposition is contracted with the determiner, that is, in the PP "\textit{no Médio Oriente}", we observe the contraction of "\textit{em}" ("\textit{in}") + "\textit{o}" ("\textit{the}"). The event argument in Portuguese is "\textit{o Médio Oriente}" and not "\textit{no Médio Oriente}". While manually annotating the test set of the ACE-2005 Portuguese version, our expert followed ACE-2005  guidelines, which meant separating the preposition from the determiner. However, this requirement poses a challenge for automatic alignment.


We detected 242 instances where our pipeline failed to correctly align the determiner, accounting for 69.14\% of the errors in argument alignment. This is one of the main reasons for the discrepancy between the relaxed (87.77\%) and Exact (70.55\%) scores. By using the exact match, these cases are considered completely misaligned. On the other hand, by using the relaxed match, despite being penalised for omitting the determiner, the pipeline still receives credit for identifying the remaining correctly aligned tokens.


We also conducted an error analysis on the pipeline text alignment components. For instance, character-level similarity matching disregards the order of characters and lacks semantic understanding. In the previous example, the argument annotation "We" was translated to "\textit{Nós}" in isolation. By relying on character similarity matching, "\textit{Nós}" was mistakenly aligned with the word "\textit{no}" (the) due to their shared characters, resulting in an incorrect alignment. This issue becomes more prevalent when dealing with shorter strings, thus, this method was not used to align triggers which are usually composed of one single word. On the other hand, the lemmatization method also introduces ambiguity, as multiple words with different meanings can map to the same lemma. This ambiguity may result in incorrect alignments, where annotations are aligned with incorrect spans in the translated text due to multiple possible lemma matches. 

In Algorithm \ref{alg:align_wordaligner}, we determine the translated annotation offsets using the maximum and minimum indexes of the previously computed alignments. However, in case of alignment errors, the calculated minimum or maximum values can become disproportionately large or small, resulting in absurdly large annotation spans. To address this issue, we implemented a safeguard by comparing the lengths of the aligned annotation and the source annotation. If a significant discrepancy is detected, the alignment candidate is discarded.

\section{Conclusion}

In this work, we developed a translation and alignment pipeline specifically designed to automatically translate the ACE-2005 corpus into both European and Brazilian Portuguese. To assess its effectiveness, we compared the pipeline results against manually aligned annotations performed by a linguistic expert, achieving a relaxed score of 87.77\% and an exact match of 70.55\%. While our primary focus was the Portuguese language, our pipeline can be easily adapted to accommodate other languages and corpora. This work expanded the usability of the ACE-2005 corpus beyond its original three languages by creating a Portuguese version, which has been accepted for publication by the LDC.

In the future, our pipeline could translate ACE-2005 into other languages and extend to other corpora such as the Entities, Relations and Events corpus (ERE) \cite{AB2/7KH7V4_2023},  enhancing NLP tasks in various languages and tasks.

\section{Acknowledgments}


This work is financed by National Funds through the Portuguese funding agency, FCT - Fundação para a Ciência e a Tecnologia, within project LA/P/0063/2020. DOI 10.54499/LA/P/0063/2020 | https://doi.org/10.54499/LA/P/0063/2020. The author also would like to acknowledge the project StorySense, with reference 2022.093 12.PTDC (DOI 10.54499/2022.09312.PTDC).

\bibliographystyle{ACM-Reference-Format}
\bibliography{sample-base}


\begin{thebibliography}{21}


\ifx \showCODEN    \undefined \def \showCODEN     #1{\unskip}     \fi
\ifx \showDOI      \undefined \def \showDOI       #1{#1}\fi
\ifx \showISBNx    \undefined \def \showISBNx     #1{\unskip}     \fi
\ifx \showISBNxiii \undefined \def \showISBNxiii  #1{\unskip}     \fi
\ifx \showISSN     \undefined \def \showISSN      #1{\unskip}     \fi
\ifx \showLCCN     \undefined \def \showLCCN      #1{\unskip}     \fi
\ifx \shownote     \undefined \def \shownote      #1{#1}          \fi
\ifx \showarticletitle \undefined \def \showarticletitle #1{#1}   \fi
\ifx \showURL      \undefined \def \showURL       {\relax}        \fi
\providecommand\bibfield[2]{#2}
\providecommand\bibinfo[2]{#2}
\providecommand\natexlab[1]{#1}
\providecommand\showeprint[2][]{arXiv:#2}

\bibitem[ace(2005)]%
        {aceguidelines}
 \bibinfo{year}{2005}\natexlab{}.
\newblock \showarticletitle{English Annotation Guidelines for Events}.
\newblock \bibinfo{journal}{\emph{Linguistic Data Consortium}} (\bibinfo{year}{2005}).
\newblock
\urldef\tempurl%
\url{https://www.ldc.upenn.edu/sites/www.ldc.upenn.edu/files/english-events-guidelines-v5.4.3.pdf}
\showURL{%
\tempurl}


\bibitem[Balali et~al\mbox{.}(2020)]%
        {BALALI2020106492}
\bibfield{author}{\bibinfo{person}{Ali Balali}, \bibinfo{person}{Masoud Asadpour}, \bibinfo{person}{Ricardo Campos}, {and} \bibinfo{person}{Adam Jatowt}.} \bibinfo{year}{2020}\natexlab{}.
\newblock \showarticletitle{Joint event extraction along shortest dependency paths using graph convolutional networks}.
\newblock \bibinfo{journal}{\emph{Knowledge-Based Systems}}  \bibinfo{volume}{210} (\bibinfo{year}{2020}), \bibinfo{pages}{106492}.
\newblock
\showISSN{0950-7051}
\urldef\tempurl%
\url{https://doi.org/10.1016/j.knosys.2020.106492}
\showDOI{\tempurl}


\bibitem[Barbosa et~al\mbox{.}(2005)]%
        {barbosa2005null}
\bibfield{author}{\bibinfo{person}{Pilar Barbosa}, \bibinfo{person}{Maria Eug{\^e}nia~L Duarte}, {and} \bibinfo{person}{Mary~Aizawa Kato}.} \bibinfo{year}{2005}\natexlab{}.
\newblock \showarticletitle{Null subjects in European and Brazilian Portuguese}.
\newblock \bibinfo{journal}{\emph{Journal of Portuguese Linguistics}} \bibinfo{volume}{4}, \bibinfo{number}{2} (\bibinfo{year}{2005}).
\newblock


\bibitem[Carrino et~al\mbox{.}(2020)]%
        {Carrino2020}
\bibfield{author}{\bibinfo{person}{Casimiro~Pio Carrino}, \bibinfo{person}{Marta~R Costa-Jussà}, {and} \bibinfo{person}{José A~R Fonollosa}.} \bibinfo{year}{2020}\natexlab{}.
\newblock \showarticletitle{Automatic Spanish Translation of the SQuAD Dataset for Multilingual Question Answering}.
\newblock  (\bibinfo{year}{2020}), \bibinfo{pages}{11--16}.
\newblock


\bibitem[Chen et~al\mbox{.}(2023)]%
        {AB2/7KH7V4_2023}
\bibfield{author}{\bibinfo{person}{Song Chen}, \bibinfo{person}{Ann Bies}, \bibinfo{person}{Kira Griffitt}, \bibinfo{person}{Joe Ellis}, {and} \bibinfo{person}{Stephanie Strassel}.} \bibinfo{year}{2023}\natexlab{}.
\newblock \bibinfo{title}{{DEFT English Light and Rich ERE Annotation}}.
\newblock
\newblock
\urldef\tempurl%
\url{https://doi.org/11272.1/AB2/7KH7V4}
\showDOI{\tempurl}


\bibitem[Cunha et~al\mbox{.}(2023)]%
        {cunha2023}
\bibfield{author}{\bibinfo{person}{Luís~Filipe Cunha}, \bibinfo{person}{Ricardo Campos}, {and} \bibinfo{person}{Alípio Jorge}.} \bibinfo{year}{2023}\natexlab{}.
\newblock \showarticletitle{Event Extraction for Portuguese: A QA-driven Approach using ACE-2005}. In \bibinfo{booktitle}{\emph{Springer’s LNAI – Lecture Notes in Artificial Intelligence}}. \bibinfo{publisher}{Springer}.
\newblock


\bibitem[Daniel and Martin(2023)]%
        {jurafsky:2023}
\bibfield{author}{\bibinfo{person}{Jurafsky Daniel} {and} \bibinfo{person}{James~H Martin}.} \bibinfo{year}{2023}\natexlab{}.
\newblock \showarticletitle{Relation and Event Extraction}.
\newblock In \bibinfo{booktitle}{\emph{Speech and Language Processing}}. Chapter~21, \bibinfo{pages}{429--445}.
\newblock


\bibitem[Devlin et~al\mbox{.}(2018)]%
        {DBLP:journals/corr/abs-1810-04805}
\bibfield{author}{\bibinfo{person}{Jacob Devlin}, \bibinfo{person}{Ming{-}Wei Chang}, \bibinfo{person}{Kenton Lee}, {and} \bibinfo{person}{Kristina Toutanova}.} \bibinfo{year}{2018}\natexlab{}.
\newblock \showarticletitle{{BERT:} Pre-training of Deep Bidirectional Transformers for Language Understanding}.
\newblock \bibinfo{journal}{\emph{CoRR}}  \bibinfo{volume}{abs/1810.04805} (\bibinfo{year}{2018}).
\newblock
\showeprint[arxiv]{1810.04805}


\bibitem[Doddington et~al\mbox{.}(2004)]%
        {doddington-etal-2004-automatic}
\bibfield{author}{\bibinfo{person}{George Doddington}, \bibinfo{person}{Alexis Mitchell}, \bibinfo{person}{Mark Przybocki}, \bibinfo{person}{Lance Ramshaw}, \bibinfo{person}{Stephanie Strassel}, {and} \bibinfo{person}{Ralph Weischedel}.} \bibinfo{year}{2004}\natexlab{}.
\newblock \showarticletitle{The Automatic Content Extraction ({ACE}) Program {--} Tasks, Data, and Evaluation}. In \bibinfo{booktitle}{\emph{Proceedings of the Fourth International Conference on Language Resources and Evaluation ({LREC}{'}04)}}. \bibinfo{publisher}{European Language Resources Association (ELRA)}, \bibinfo{address}{Lisbon, Portugal}.
\newblock


\bibitem[Dou and Neubig(2021)]%
        {Dou2021}
\bibfield{author}{\bibinfo{person}{Zi~Yi Dou} {and} \bibinfo{person}{Graham Neubig}.} \bibinfo{year}{2021}\natexlab{}.
\newblock \showarticletitle{Word Alignment by Fine-tuning Embeddings on Parallel Corpora}.
\newblock \bibinfo{journal}{\emph{EACL 2021 - 16th Conference of the European Chapter of the Association for Computational Linguistics, Proceedings of the Conference}} (\bibinfo{year}{2021}), \bibinfo{pages}{2112--2128}.
\newblock


\bibitem[Du and Cardie(2020)]%
        {Du2020}
\bibfield{author}{\bibinfo{person}{Xinya Du} {and} \bibinfo{person}{Claire Cardie}.} \bibinfo{year}{2020}\natexlab{}.
\newblock \showarticletitle{Event Extraction by Answering (Almost) Natural Questions}.
\newblock \bibinfo{journal}{\emph{EMNLP 2020 - 2020 Conference on Empirical Methods in Natural Language Processing, Proceedings of the Conference}} (\bibinfo{year}{2020}), \bibinfo{pages}{671--683}.
\newblock
\showISBNx{9781952148606}


\bibitem[Levenshtein(1965)]%
        {levenshtein1965}
\bibfield{author}{\bibinfo{person}{Vladimir~I Levenshtein}.} \bibinfo{year}{1965}\natexlab{}.
\newblock \showarticletitle{Binary Codes Capable of Correcting Deletions, Insertions, and Reversals}.
\newblock  \bibinfo{volume}{163}, \bibinfo{number}{4} (\bibinfo{year}{1965}), \bibinfo{pages}{845--848}.
\newblock


\bibitem[Li et~al\mbox{.}(2013)]%
        {li-etal-2013-joint}
\bibfield{author}{\bibinfo{person}{Qi Li}, \bibinfo{person}{Heng Ji}, {and} \bibinfo{person}{Liang Huang}.} \bibinfo{year}{2013}\natexlab{}.
\newblock \showarticletitle{Joint Event Extraction via Structured Prediction with Global Features}. In \bibinfo{booktitle}{\emph{Proceedings of the 51st Annual Meeting of the Association for Computational Linguistics (Volume 1: Long Papers)}}. \bibinfo{address}{Bulgaria}, \bibinfo{pages}{73--82}.
\newblock


\bibitem[Li et~al\mbox{.}(2022)]%
        {li2022survey}
\bibfield{author}{\bibinfo{person}{Qian Li}, \bibinfo{person}{Jianxin Li}, \bibinfo{person}{Jiawei Sheng}, \bibinfo{person}{Shiyao Cui}, \bibinfo{person}{Jia Wu}, \bibinfo{person}{Yiming Hei}, \bibinfo{person}{Hao Peng}, \bibinfo{person}{Shu Guo}, \bibinfo{person}{Lihong Wang}, \bibinfo{person}{Amin Beheshti}, {and} \bibinfo{person}{Philip~S. Yu}.} \bibinfo{year}{2022}\natexlab{}.
\newblock \bibinfo{title}{A Survey on Deep Learning Event Extraction: Approaches and Applications}.
\newblock
\newblock
\showeprint[arxiv]{2107.02126}~[cs.CL]


\bibitem[Lin et~al\mbox{.}(2020)]%
        {lin-etal-2020-joint}
\bibfield{author}{\bibinfo{person}{Ying Lin}, \bibinfo{person}{Heng Ji}, \bibinfo{person}{Fei Huang}, {and} \bibinfo{person}{Lingfei Wu}.} \bibinfo{year}{2020}\natexlab{}.
\newblock \showarticletitle{A Joint Neural Model for Information Extraction with Global Features}. In \bibinfo{booktitle}{\emph{Proceedings of the 58th Annual Meeting of the Association for Computational Linguistics}}. \bibinfo{pages}{7999--8009}.
\newblock


\bibitem[Lu et~al\mbox{.}(2021)]%
        {lu-etal-2021-text2event}
\bibfield{author}{\bibinfo{person}{Yaojie Lu}, \bibinfo{person}{Hongyu Lin}, \bibinfo{person}{Jin Xu}, \bibinfo{person}{Xianpei Han}, \bibinfo{person}{Jialong Tang}, \bibinfo{person}{Annan Li}, \bibinfo{person}{Le Sun}, \bibinfo{person}{Meng Liao}, {and} \bibinfo{person}{Shaoyi Chen}.} \bibinfo{year}{2021}\natexlab{}.
\newblock \showarticletitle{{T}ext2{E}vent: Controllable Sequence-to-Structure Generation for End-to-end Event Extraction}. In \bibinfo{booktitle}{\emph{Proceedings of the 59th Annual Meeting of the Association for Computational Linguistics and the 11th International Joint Conference on Natural Language Processing (Volume 1: Long Papers)}}. \bibinfo{pages}{2795--2806}.
\newblock


\bibitem[Nguyen et~al\mbox{.}(2021)]%
        {nguyen-etal-2021-cross}
\bibfield{author}{\bibinfo{person}{Minh~Van Nguyen}, \bibinfo{person}{Viet~Dac Lai}, {and} \bibinfo{person}{Thien~Huu Nguyen}.} \bibinfo{year}{2021}\natexlab{}.
\newblock \showarticletitle{Cross-Task Instance Representation Interactions and Label Dependencies for Joint Information Extraction with Graph Convolutional Networks}. In \bibinfo{booktitle}{\emph{Proceedings of the 2021 Conference of the North American Chapter of the Association for Computational Linguistics: Human Language Technologies}}. \bibinfo{pages}{27--38}.
\newblock
\urldef\tempurl%
\url{https://doi.org/10.18653/v1/2021.naacl-main.3}
\showDOI{\tempurl}


\bibitem[Nguyen et~al\mbox{.}(2022)]%
        {nguyen-etal-2022-joint}
\bibfield{author}{\bibinfo{person}{Minh~Van Nguyen}, \bibinfo{person}{Bonan Min}, \bibinfo{person}{Franck Dernoncourt}, {and} \bibinfo{person}{Thien Nguyen}.} \bibinfo{year}{2022}\natexlab{}.
\newblock \showarticletitle{Joint Extraction of Entities, Relations, and Events via Modeling Inter-Instance and Inter-Label Dependencies}. In \bibinfo{booktitle}{\emph{Proceedings of the 2022 Conference of the North American Chapter of the Association for Computational Linguistics: Human Language Technologies}}. \bibinfo{address}{United States}, \bibinfo{pages}{4363--4374}.
\newblock
\urldef\tempurl%
\url{https://doi.org/10.18653/v1/2022.naacl-main.324}
\showDOI{\tempurl}


\bibitem[Nguyen et~al\mbox{.}(2016)]%
        {nguyen-etal-2016-joint-event}
\bibfield{author}{\bibinfo{person}{Thien~Huu Nguyen}, \bibinfo{person}{Kyunghyun Cho}, {and} \bibinfo{person}{Ralph Grishman}.} \bibinfo{year}{2016}\natexlab{}.
\newblock \showarticletitle{Joint Event Extraction via Recurrent Neural Networks}. In \bibinfo{booktitle}{\emph{Proceedings of the 2016 Conference of the North {A}merican Chapter of the Association for Computational Linguistics: Human Language Technologies}}. \bibinfo{address}{California}, \bibinfo{pages}{300--309}.
\newblock
\urldef\tempurl%
\url{https://doi.org/10.18653/v1/N16-1034}
\showDOI{\tempurl}


\bibitem[Ratcliff and Metzener(1988)]%
        {ratcliff1988pattern}
\bibfield{author}{\bibinfo{person}{John~W Ratcliff} {and} \bibinfo{person}{David~E Metzener}.} \bibinfo{year}{1988}\natexlab{}.
\newblock \showarticletitle{Pattern-matching-the gestalt approach}.
\newblock \bibinfo{journal}{\emph{Dr Dobbs Journal}} \bibinfo{volume}{13}, \bibinfo{number}{7} (\bibinfo{year}{1988}), \bibinfo{pages}{46}.
\newblock


\bibitem[Vaswani et~al\mbox{.}(2017)]%
        {vaswani2017attention}
\bibfield{author}{\bibinfo{person}{Ashish Vaswani}, \bibinfo{person}{Noam Shazeer}, \bibinfo{person}{Niki Parmar}, \bibinfo{person}{Jakob Uszkoreit}, \bibinfo{person}{Llion Jones}, \bibinfo{person}{Aidan~N. Gomez}, \bibinfo{person}{Lukasz Kaiser}, {and} \bibinfo{person}{Illia Polosukhin}.} \bibinfo{year}{2017}\natexlab{}.
\newblock \bibinfo{title}{Attention Is All You Need}.
\newblock
\newblock
\showeprint[arxiv]{1706.03762}~[cs.CL]


\end{thebibliography}

\end{document}